\documentclass[conference]{IEEEtran}
\input epsf
\usepackage{graphicx}
\usepackage[sorting=none]{biblatex}
\usepackage{amsmath,amssymb}
\usepackage{algorithm}
\usepackage{algorithmicx}
\usepackage{algpseudocode}

\begin{document}

\title{\LARGE VLLFL: A Vision-Language Model Based Lightweight Federated Learning Framework for Smart Agriculture }

\author{\IEEEauthorblockN{Long Li\IEEEauthorrefmark{1},Jiajia Li\IEEEauthorrefmark{2}, Dong Chen\IEEEauthorrefmark{3}, Lina Pu\IEEEauthorrefmark{4}, Haibo Yao\IEEEauthorrefmark{5}, *Yanbo Huang\IEEEauthorrefmark{5} }
\\
\IEEEauthorblockA{\IEEEauthorrefmark{1} Department of Electrical and Computer Engineering, The University of Alabama \\
                    \IEEEauthorrefmark{2} Electrical and Computer Engineering, Michigan State University \\
                     \IEEEauthorrefmark{3} Agricultural and Biological Engineering, Mississippi State University \\
                     \IEEEauthorrefmark{4} Department of Computer Science, University of Alabama \\
                     \IEEEauthorrefmark{5} USDA-ARS Genetics and Sustainbale Agriculture \\
}
}

\maketitle
\renewcommand{\thefootnote}{} 
\footnotetext{* Corresponding author : yanbo.huang@usda.gov}
\addtocounter{footnote}{-1}   

\begin{abstract}
In modern smart agriculture, object detection plays a crucial role by enabling automation, precision farming, and monitoring of resources. From identifying crop health and pest infestations to optimizing harvesting processes, accurate object detection enhances both productivity and sustainability. However, training object detection models often requires large-scale data collection and raises privacy concerns, particularly when sensitive agricultural data is distributed across farms. To address these challenges, we propose VLLFL, a vision-language model-based lightweight federated learning framework (VLLFL). It harnesses the generalization and context-aware detection capabilities of the vision-language model (VLM) and leverages the privacy-preserving nature of federated learning. By training a compact prompt generator to boost the performance of the VLM deployed across different farms, VLLFL preserves privacy while reducing communication overhead. Experimental results demonstrate that VLLFL achieves 14.53\% improvement in the performance of VLM while reducing 99.3\% communication overhead. Spanning tasks from identifying a wide variety of fruits to detecting harmful animals in agriculture, the proposed framework offers an efficient, scalable, and privacy-preserving solution specifically tailored to agricultural applications.
\end{abstract}
\IEEEoverridecommandlockouts

%
\IEEEpeerreviewmaketitle

\section{Introduction}
In recent years, smart agriculture has emerged as a transformative approach to increase farming efficiency, reduce costs, and maintain environmental sustainability \cite{ghazal2024computer, li2022reinforcement}. By incorporating cutting-edge technologies such as the Internet of Things (IoT), artificial intelligence (AI), and advanced data analytics, smart agriculture offers improved monitoring and decision-making across the entire agricultural supply chain \cite{esfandiari2025multi}. One crucial component of many smart farming solutions is object detection, which enables systems to identify crops, weeds, pests, and machinery in real-time through sensor networks or robotic platforms \cite{akbar2024comprehensive}. This real-time detection capability is instrumental in tasks like early pest detection \cite{lippi2022data} and precise harvesting \cite{zhang2020multi}, all of which can significantly enhance yield and utilization of resource \cite{li2024foundation}. 

Traditional object detection techniques, typically based on convolutional neural networks (CNNs) like Faster R-CNN \cite{zhang2024deep} or YOLO \cite{li2021q}, have demonstrated considerable success in classifying and localizing objects.\cite{gugssa2023enhancing}. These models typically rely on visual features—such as color, texture, and shape—and perform effectively when trained on sufficiently large and representative datasets \cite{lin2024enhancing}. However, their performance can be limited in complex agricultural environments where visually similar objects might have different contextual meanings (e.g., healthy leaves versus early-stage blight) and annotations are costly to obtain. By contrast, vision-language models integrate textual and visual information, enabling them to capture more semantic details than purely visual models \cite{liu2024grounding,tran2023ai,li2023knowledge}. In agricultural tasks, for example, vision-language models can interpret textual descriptions of symptoms or growth stages, providing richer insights and more accurate identification of diseases, nutrient deficiencies, or weed species \cite{8961482}. Consequently, these models offer a powerful alternative, enhancing interpretability and reducing misclassification by combining linguistic cues with visual signals \cite{zhang2024vision}. While traditional object detection methods remain integral for many baseline tasks, the added semantic understanding of vision-language approaches increasingly positions them as a more comprehensive solution for intelligent farming systems \cite{zhu2024harnessing,liu2024multimodal}. 

Despite the effectiveness of both traditional and vision-language object detection methods, large-scale data collection is often required to ensure robust performance across diverse agricultural environments \cite{li2024secure}. However, assembling such extensive datasets is challenging. In particular, those containing high-resolution farmland images and potentially identifiable individuals raise significant privacy concerns \cite{li2025extra}. These concerns include meeting requirements set forth by regulations such as the General Data Protection Regulation (GDPR) \cite{gupta2020security}. Federated learning (FL) has emerged as a promising approach to address these issues by training a global model across multiple decentralized clients without requiring the direct exchange of raw data. Instead, each client conducts local training and only shares updated parameters, preserving data confidentiality and reducing the risk of unauthorized data exposure \cite{li2020federated, gaur2024impact}. 

However, standard federated learning solutions introduce another obstacle: communication overhead \cite{al2021reducing}. In a typical FL setup, models—often quite large in object detection or vision-language tasks—must be frequently synchronized between clients and a central server. This repeated transmission of model weights can be bandwidth-intensive, especially in agricultural environments where network connections may be limited or costly \cite{zhang2024vision}.For example, a typical object detection model like YOLOv3 includes approximately 62 million trainable parameters, totaling 1,984 Mb. At a transmission speed of 100 Mbps, it takes around 19 seconds for each client to upload the model to the server in each round of federated learning. As a result, practitioners often face a trade-off between privacy protection and practical efficiency; they might reduce synchronization frequency to save bandwidth at the risk of slower convergence or degraded accuracy. 

\begin{figure}[htp]
    \centering
    \includegraphics[width=0.9\linewidth]{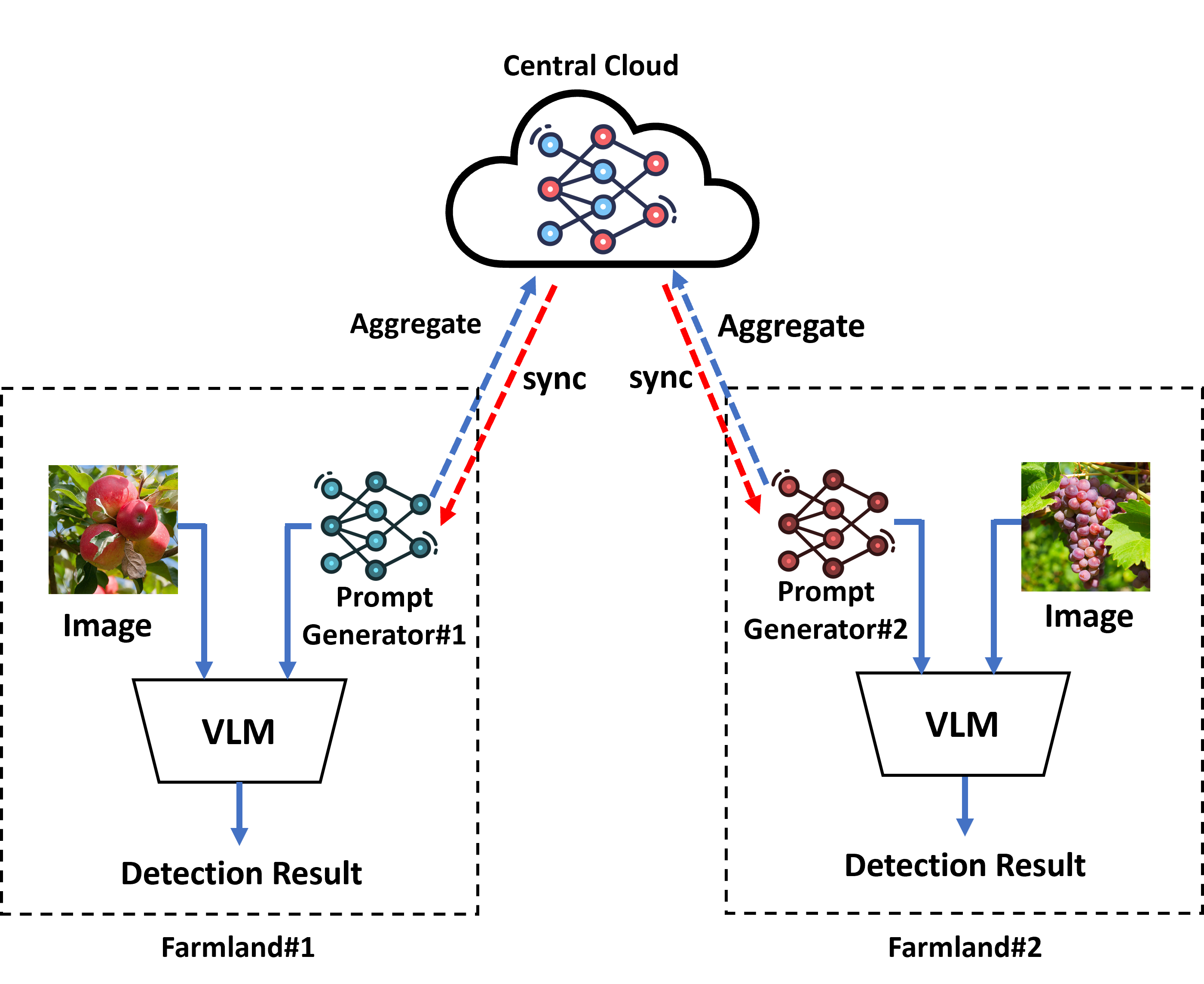}
    \caption{Application structure of VLLFL in smart agriculture}
    \label{fig:scenario}
\end{figure}

To address the aforementioned challenges, we propose a Vision-Language model-based Lightweight Federated Learning framework (VLLFL) that integrates the semantic strengths of multimodal approaches with the privacy-preserving benefits of federated learning, while simultaneously minimizing communication overhead. An example structure of VLLFL in smart agriculture is presented in Fig. \ref{fig:scenario}. Inspired by the design philosophy presented in \cite{qiu2023text}, which demonstrated the feasibility of training prompt generators in a federated manner for classification tasks, our approach focuses on learning a compact prompt generator that interacts with the model’s text encoder. This strategy significantly reduces the number of parameters that must be exchanged, thereby lowering bandwidth consumption and accelerating convergence. In this way, VLLFL preserves the contextual understanding of vision-language models for object detection, ensures stringent data privacy through federated learning, and enables an efficient model update process suitable for large-scale agricultural deployments. In our experiments, VLLFL significantly enhanced the object detection accuracy of the VLM for fruit detection across multiple classes and clients. Specifically, VLLFL achieved an average improvement in mean average precision (mAP) from 9.59\% to 24.12\% while reduce communication overhead by 99.3\%, thereby demonstrating VLLFL’s superior efficiency in deploying vision-language models via federated learning.

The contribution of this work are as below:
\begin{enumerate}
    \item We apply federated learning to integrate a large vision-language model, GroundingDINO, into smart agriculture applications, thereby enhancing object detection capabilities while preserving the privacy of locally stored data. 
    \item We introduce a novel framework—VLLFL—that significantly reduces the communication overhead when deploying vision language model in federated learning. By training only a lightweight prompt generator instead of the entire model, our approach maintains high detection accuracy while minimizing bandwidth and computational costs. 
    \item We conduct extensive experiments to evaluate the effectiveness and efficiency of the proposed framework. Results show that VLLFL significantly enhances VLM’s performance across various object detection tasks, with minimal communication overhead—demonstrating its superior suitability for deploying vision-language models in federated learning settings. 
\end{enumerate}

The remainder of this paper is organized as follows: Section II provides the necessary background knowledge. Section III presents the detailed design of the proposed method. Section IV reports the comprehensive evaluation results and discussion. Finally, Section V concludes the paper.

\section{Background}

\subsection{Vision Language Model}
Vision-language models have become increasingly popular for their ability to jointly process textual and visual content, unlocking more context-aware performance than single-modality architectures \cite{li2024vision}. By integrating linguistic cues—such as object labels or domain-specific descriptions—with visual features, these models capture semantic nuances often overlooked by traditional object detection approaches, which typically rely on attributes like color, texture, and shape. For example, while a conventional detector might classify an object simply as an “apple” based on visual characteristics, a vision-language model can further distinguish between varieties such as “green apple” and “red apple” if such labels are provided. This cross-modal capability not only enhances detection precision but also facilitates a deeper interpretation of objects in real-world environments—particularly valuable in agriculture, where visually similar varieties often demand fine-grained differentiation \cite{ghosh2024exploring}.

In recent years, numerous widely adopted vision-language models have demonstrated their versatility across various domains, including agriculture. Early approaches, such as the two-tower architecture proposed by \cite{kiros2014unifying}, aligned separate image and text embeddings within a shared space, enabling basic cross-modal retrieval. More advanced transformer-based models—including ViLBERT \cite{lu2019vilbert}, LXMERT \cite{tan2019lxmert}, and UNITER \cite{chen2020uniter}—introduced multi-layer cross-attention mechanisms, significantly improving joint image-text representations. Trained on large-scale datasets, these models achieved strong performance in tasks such as visual question answering and referring expression comprehension. Subsequently, CLIP \cite{radford2021learning} achieved impressive zero-shot classification by learning generalized representations from large-scale image-text corpora, while DALL·E \cite{ramesh2021zero} expanded the frontier by generating images from textual prompts. In agricultural contexts, such models offer promising capabilities for detecting pests, identifying crop diseases, or generating descriptive captions of field conditions—providing farmers with detailed, context-rich insights to support informed decision-making. 

Among vision-language paradigms, GroundingDINO \cite{liu2024grounding} stands out for its integration of object detection with language grounding. It extends DINO’s detection-oriented architecture by incorporating mechanisms that align textual prompts or labels with corresponding image regions. This design excels at generating high-quality bounding boxes guided by semantic cues—a capability particularly valuable in agricultural applications where fine-grained distinctions (e.g., between plant varieties or infection stages) are essential. As a result, we adopt GroundingDINO as the backbone of our proposed framework, leveraging its strong detection performance and linguistic grounding capabilities to advance precision agriculture.

\subsection{Federated-Learning with Vision-Language Models}

Federated learning (FL) has emerged as a powerful paradigm for enabling collaborative model training without directly sharing raw data. While it has been successfully applied to a range of machine learning tasks—such as image classification and natural language processing—its adoption in vision-language contexts is comparatively recent. The fundamental idea behind FL is to train a global model by aggregating locally computed updates (e.g., gradients or model weights) from multiple clients. This setup preserves data privacy and often complies more readily with regulations (e.g., GDPR) because the sensitive information remains on local devices or servers \cite{li2020federated}. However, deploying large-scale vision-language models within an FL framework introduces unique challenges related to communication overhead, data heterogeneity, and model complexity \cite{che2023multimodal}.

Vision-language models typically consist of high-dimensional parameters due to their attention-based architectures and large-scale pretraining objectives \cite{lu2019vilbert}. For instance, the GroundingDINO model discussed earlier contains approximately 172 million trainable parameters, in total of 656 MB. In federated learning settings, training such large models can result in substantial communication overhead, as model updates must be exchanged between clients and a central server. This issue is especially critical in domains like agriculture and healthcare, where network bandwidth is often limited and connectivity may be intermittent. Existing strategies to mitigate overhead include parameter-efficient fine-tuning methods—such as prompt tuning, adapter layers, or low-rank factorization—and gradient compression techniques that reduce the payload of each synchronization step. These approaches help make federated learning of vision-language models more feasible and cost-effective.

On the data side, federated learning for vision-language tasks often involves heterogeneous and domain-specific inputs \cite{nguyen2024flora}. In agriculture, for instance, local nodes may have imagery that corresponds to different crop varieties, growth conditions, or climate zones, alongside textual annotations that vary widely in specificity and language . This diversity can complicate the alignment of textual and visual modalities across clients, often necessitating advanced domain adaptation and continual learning strategies to ensure robust global model performance. Nonetheless, when effectively implemented, FL confers significant advantages for vision-language applications: it preserves data confidentiality, caters to multi-farm or multi-region collaborations, and leverages distributed resources for large-scale model training.

Recent frameworks have shown promise in bridging vision-language models and federated learning. Some efforts focus on specialized tasks like caption generation or referring expression comprehension, relying on partial updates (e.g., prompt generators) instead of full model weights to conserve bandwidth \cite{yan2024lightweight}. Others explore knowledge distillation methods, where a centrally maintained teacher model guides local models, further reducing synchronization burden \cite{pan2024survey}. As advances in efficient training techniques and communication protocols continue, federated learning with vision-language models is poised to open up collaborative opportunities in industries where data sensitivity and distribution constraints are paramount, from smart agriculture to healthcare and beyond.

\begin{figure*}
    \centering
    \includegraphics[width=0.9\linewidth]{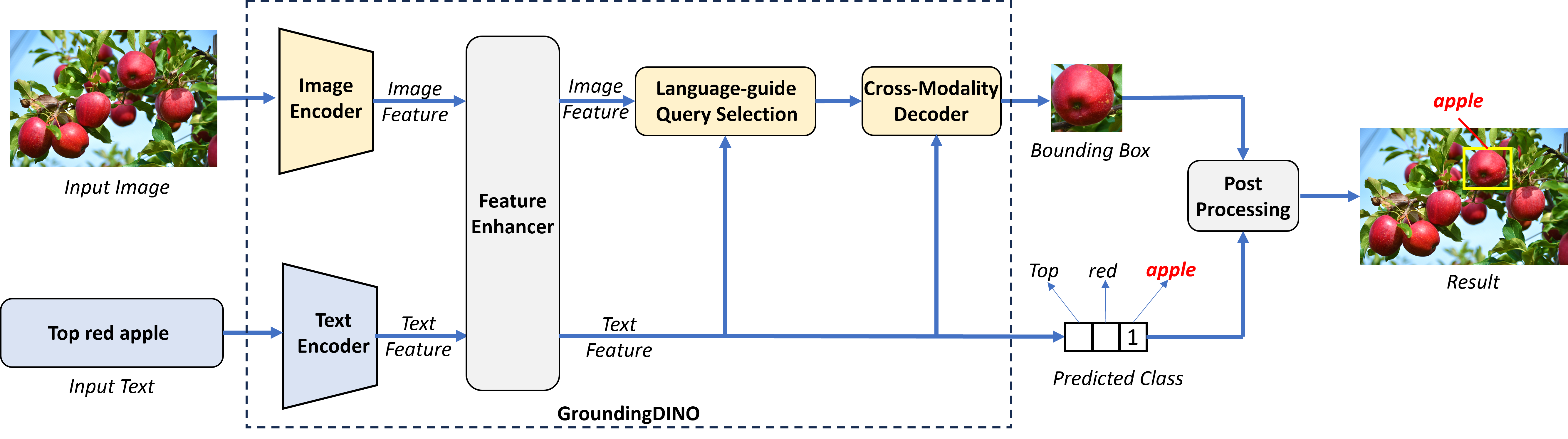}
    \caption{The architecture of VLM backbone in VLLFL based on GroundingDINO\cite{liu2024grounding}}
    \label{fig:dino}
\end{figure*}
\subsection{Vision-Language Model Prompt Learning}

Prompt learning has emerged as an effective way to adapt large vision-language models without fine-tuning the entirety of their parameters \cite{bu2024efficient}. Rather than modifying all layers, a small number of learnable “prompt tokens” or “prompt vectors” is inserted into the model, guiding its attention or output distribution toward a target task. This approach outperforms rigid, hand-crafted prompts (e.g., “a photo of a [CLASS]”) by dynamically optimizing textual cues, thus enabling more nuanced adaptations. For agriculture, prompt learning is particularly appealing because complex domain-specific vocabulary—covering different crop varieties or pest types—can be integrated without excessively retraining the underlying model \cite{li2024foundation}.

Recently, researchers have proposed a range of prompt learning methods that build on this principle. Context Optimization (CoOp) and Conditional Context Optimization (Co-CoOp) \cite{zhou2022conditional} introduce trainable prompts to large-scale vision-language models like CLIP \cite{radford2021learning}, achieving strong results in zero-shot or few-shot classification settings. Efforts like FedTPG \cite{qiu2024federated} extend prompt learning to federated learning scenarios, focusing on a lightweight prompt generator shared across multiple clients. By exchanging minimal parameter updates, FedTPG reduces communication overhead compared to conventional federated approaches. At the same time, frameworks such as PromptDet \cite{feng2022promptdet} explore prompt-based strategies for object detection; however, these methods do not currently operate in a federated context, often assuming centralized data availability for training detection models.

To date, there has been no dedicated prompt learning solution tailored specifically for federated object detection tasks. This gap is particularly notable in smart agriculture, where privacy constraints and distributed data ownership are paramount. Inspired by FedTPG’s success in federated classification with prompt learning, we propose VLLFL—a federated learning framework that leverages prompt tuning for object detection in real-world agricultural environments \cite{qiu2023text}. By focusing on training compact prompts rather than entire model weights, VLLFL seeks to preserve data privacy, minimize communication overhead, and deliver high-performing detection capabilities for tasks such as identifying crop diseases, detecting wildlife threats, or monitoring farm machinery.

\section{Method}

The proposed VLLFL framework consists of a prompt generator and a base vision-language model (VLM), and is deployed within a federated learning setting. As illustrated in Fig. \ref{fig:scenario}, the objective of VLLFL is to train a global prompt generator that facilitates the deployment of VLMs and enhances object detection performance. In this section, we present the details of the proposed VLLFL framework.

\subsection{Vision-Language Model}
Vision-language models excel in zero-shot and open-set recognition by leveraging text-based cues to detect previously unseen object categories, even without prior training examples \cite{li2024foundation}. A typical VLM architecture consists of a text encoder and an image encoder, which extract features from textual and visual inputs, respectively, and project them into a shared vector space. Predictions are then made by matching the encoded image features with their corresponding text features. This shared visual-language representation enhances robustness to domain shifts, enabling strong generalization across diverse styles and contexts. Moreover, VLMs can be rapidly adapted to a wide range of downstream tasks with minimal labeled data.

In our proposed VLLFL framework, the foundation model employed for object detection is GroundingDINO \cite{liu2024grounding}. GroundingDINO is a popular open-set detector built upon the DETR-like architecture DINO \cite{zhang2022dino}, which integrates end-to-end transformer-based detection mechanisms. The overall workflow of GroundingDINO is illustrated in Fig. \ref{fig:dino}. Initially, the model extracts visual and textual features using an image encoder (e.g., the Swin Transformer \cite{liu2021swin}) and a text encoder (e.g., BERT \cite{devlin2019bert}), respectively. These extracted features are then passed to a feature enhancer network, which fuses cross-modal information to enable comprehensive integration of textual and visual cues.

After generating enriched cross-modal representations, a language-guided query selection module selects relevant queries based on image features and corresponding text embeddings, effectively leveraging the synergy between modalities. This selection process exemplifies the model’s ability to align semantic textual descriptions with visual content, thereby enhancing detection precision and contextual awareness. The selected queries are subsequently fed into a cross-modal decoder, which refines the fused features and iteratively updates model parameters through cross-modal analysis. Finally, the decoder outputs predicted bounding boxes and associated textual phrases. During post-processing, a box threshold and a text similarity threshold are applied to filter valid predictions and extract corresponding class labels from the input text. The result is a set of accurately detected objects paired with semantically relevant class names. 

It is important to note that GroundingDINO contains 172 million trainable parameters, totaling 656 MB. Directly training such a large model in a federated learning setting would lead to substantial communication overhead. To address this, the pre-trained parameters of GroundingDINO are frozen and excluded from the training process. Instead, the VLLFL framework focuses on training a lightweight prompt generator, avoiding updates to the base model.  This design significantly reduces communication overhead during federated learning, making the framework well-suited for deployment in resource-constrained agriculture scenarios.

\subsection{Prompt Generator}

\begin{figure*}
    \centering
    \includegraphics[width=0.9\linewidth]{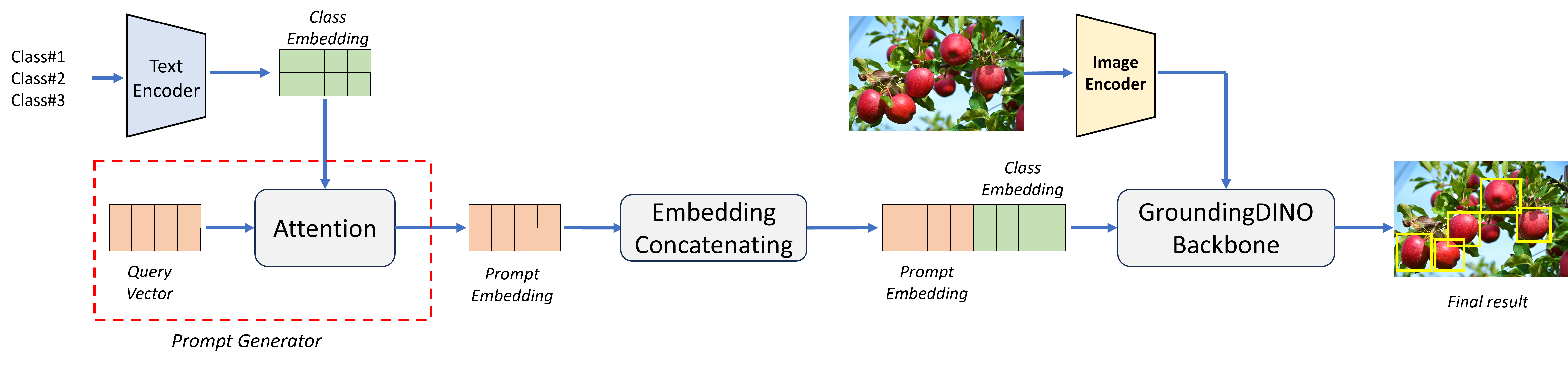}
    \caption{ Prompt generator in proposed VLLFL framework in a local client}
    \label{fig:vllfl}
\end{figure*}

In the VLLFL framework, the prompt generator is designed to provide contextual information that guides GroundingDINO in making accurate predictions. As illustrated in Fig. \ref{fig:dino}, the input text ``Top red apple" contains prompt keywords such as ``top" and ``red", which help GroundingDINO focus on the correct object. Even if the model has never encountered an ``apple" before, these descriptive cues allow it to focus on relevant visual regions—such as red objects in the top part of the image. This exemplifies the fundamental concept behind the prompt generator. We assume that effective text prompts, derived from class names, can improve object detection performance. Therefore, we train a lightweight prompt generator to produce context-aware prompt vectors, conditioned on task-specific text inputs. The structure of the VLLFL framework within a local client is shown in Fig. \ref{fig:vllfl}.

Each local client concatenates the class names associated with its specific detection task and feeds them into the text encoder of GroundingDINO. The text encoder translates this input into class embeddings, denoted as $\mathcal{T}$, which represent a set of embedding vectors corresponding to the class names. The prompt generator, denoted as $f_\theta$, then produces a set of $m$ prompt vectors $\mathcal{P}$ based on the class embeddings $\mathcal{T}$, formulated as: 

\begin{equation}
    \mathcal{P} = \{ \mathbf{v}_k \}_{k=1}^{m} = f_\theta(\mathcal{T}).
\end{equation}

The parameter $m$ represents the prompt width, which determines the number of prompt vectors generated for each class. The prompt generator $f_\theta$ is implemented as a lightweight cross-attention module composed of learnable parameters $\theta$, a query vector $Q$, and projection matrices $W_K$ and $W_V$. Given the input class embeddings $\mathcal{T}$, the prompt generator transforms contextual information into key and value vectors, denoted as $K_T$ and $V_T$, respectively. This transformation is defined by the following equations:
\begin{align}
    K_\mathcal{T} &= \mathcal{T} \times W_K, \\
    V_\mathcal{T} &= \mathcal{T} \times W_V.
\end{align}

The cross-attention layer integrates the key and value vectors with the learnable query vector $Q$. The output of this layer is then passed through hidden layers $h_\phi$, to produce the final prompt vectors $\mathcal{P}$. This process is defined as follows:
\begin{equation}
    f_\theta(\mathcal{T}) = h_\phi(\text{CrossAttention}(Q, K_\mathcal{T}, V_\mathcal{T}))
\end{equation}

As illustrated in Fig. \ref{fig:vllfl}, the generated prompts reside directly in the embedding space. Unlike traditional prompt engineering in the text space—where prompts are interpretable by humans—VLLFL operates entirely in the embedding space to maximize performance. Interpretability is not a design objective of VLLFL, as it may compromise computational efficiency. After generating the prompt vectors $\mathcal{P}$, these vectors are concatenated with the original class embeddings $\mathcal{T}$ and fed into GroundingDINO as the new text input for feature extraction. Combined with the image features, GroundingDINO then produces the final prediction results.

\begin{figure}
    \centering
    \includegraphics[width=0.7\linewidth]{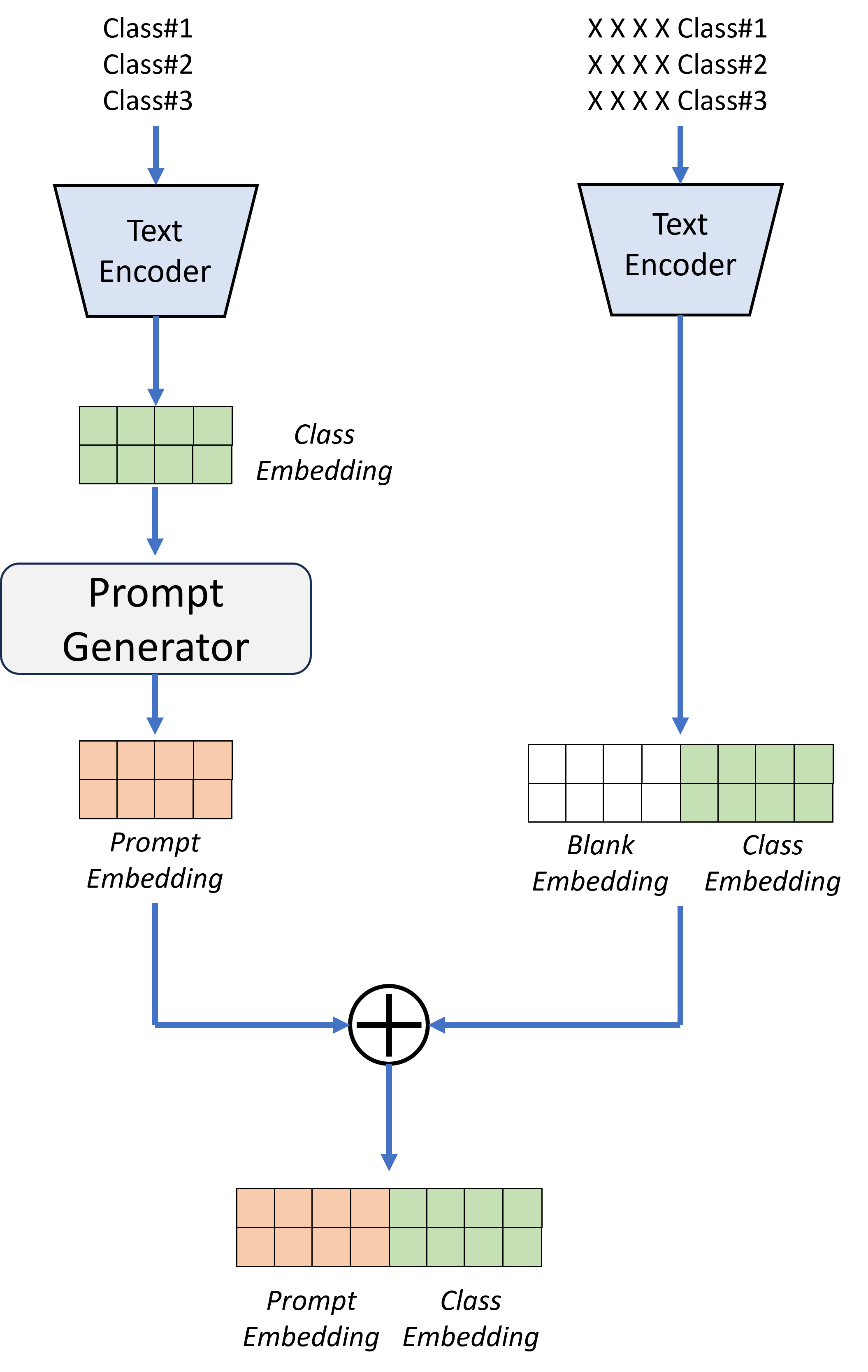}
    \caption{ Embedding concatenating in VLLFL}
    \label{fig:concate}
\end{figure}

However, directly concatenating prompt and class embeddings without structural alignment may distort the original embedding space and reduce effectiveness. To address this, VLLFL employs a structured embedding alignment process, as shown in Fig. \ref{fig:concate}.

Specifically, the prompt generator creates prompt embeddings based on the input class embeddings. In parallel, a new textual input is formed by prefixing each class name with a sequence of placeholder tokens (e.g., ``X"), where the number of placeholders equals the prompt width $m$. When this input is passed through the text encoder, it produces a composite embedding consisting of the original class embeddings and a set of meaningless placeholder embeddings. We then replace these placeholder embeddings with the generated prompt vectors $\mathcal{P}$, resulting in the final prompted embeddings used for downstream detection.

In the VLLFL framework, the prompt generator is lightweight, comprising only about 1 million trainable parameters—equivalent to 3.81 MB. During each round of federated learning, only this small set of parameters needs to be transmitted by each client, reducing communication overhead by approximately 99.3\% compared to directly training the full base vision-language model.

\subsection{Federated-learning Model Aggregation}
The proposed VLLFL operates in a federated learning in FL settings, where multiple remote clients, each handling distinct object detection tasks, collaboratively train a shared prompt generator. The training procedure of VLLFL is outlined in Algorithm \ref{alg:vllfl}. We assume that each client possesses a unique dataset corresponding to its specific detection task. Initially, the central server initializes the global prompt generator $f_{\theta}$ with parameters $\theta^0$. In each communication round, a random subset of clients, denoted by $S^r$, is selected to participate. These clients perform local training on their respective datasets and transmit the updated model parameters back to the server. The server then aggregates these updates to refine the global prompt generator.

The full training process of the VLLFL framework is described as follows.

\begin{itemize}
    \item \textbf{Step I:} The server selects a random subset of remote clients, denoted as $S^r$, to participate in the current round of federated training. Each selected client $i \in S^r$ receives the latest global parameters $\theta^r$ to configure its local prompt generator $f_{\theta^r}$.

    \item \textbf{Step II:} Each client generates class name embeddings $T_i$ by inputting its local detection class names into the GroundingDINO text encoder. The prompt generator $f_{\theta^r}$ is then trained locally for $K$ epochs using the client’s private dataset.

    \item \textbf{Step III:} After completing local training, each client sends its updated prompt generator parameters $\theta_i^{r+1}$ back to the server.

    \item \textbf{Step IV:} The server aggregates the received parameters to update the global model via $\theta^{r+1} \gets \frac{1}{|S^r|} \sum_{i \in S^r} \theta_i^{r+1}$. The updated parameters are then distributed for the next round, and the process repeats from Step I.
\end{itemize}

\begin{algorithm}[ht]
\caption{VLLFL Algorithm \cite{qiu2024federated}}
\label{alg:vllfl}
\begin{algorithmic}[1]  

\Require Number of communication rounds $R$, number of local epochs $K$, 
         initialization parameters $\theta^0$ 

\vspace{0.5em}
\State \textbf{Server executes:}
\State Initialize prompt generator $f_{\theta}$ with parameters $\theta^0$.
\For{$r \gets 0$ to $R$}
  \State Pick a random subset of remote clients as $S^r$.
  \For{$i \in S^r$}
    \State Receive the local model parameters $\theta_{i}^{r+1}$ 
           from \textit{Client $i$}.
  \EndFor
  \State Aggregate the model parameters:
  \State $\theta^{r+1} \gets \frac{1}{|S^r|} \sum_{i \in S^r} \theta_i^{r+1}$
  \State Distribute the current global model $\theta_i^{r+1}$ to all selected clients.
\EndFor

\noindent\hrulefill
\vspace{0.5em}

\State
\textbf{Executed by client $i$:}
\State Obtain the final model parameter $\theta^r$ from server.
\State Obtain the class name embeddings $T_i$

\For{$k \gets 0$ to $K$}
  \State Generate the prompt vectors:
  \State $P_i^r \gets f_{\theta}(T_i)$
  \State Concatenate the prompt vectors with class embedding
  \State Perform forward inference and get prediction result 
  \State Update parameters from $\theta^r$ to $\theta_i^{r+1}$ locally.
\EndFor

\end{algorithmic}
\end{algorithm}

Through the federated learning framework, diverse contextual and semantic information from multiple remote clients—each handling different object detection tasks—contributes to improved model generalization. By encoding task-specific text embeddings, each client enriches the global prompt generator with unique contextual cues. As a result, the collaboratively trained model is capable of serving a wide range of detection scenarios without overfitting to any single task.

\section{Experiment}

\subsection{Experiment Setup}

\subsubsection{Dataset}

The primary dataset used in our experiments is the MetaFruit dataset \cite{li2025metafruit}, a multi-class fruit detection benchmark consisting of 4,248 high-resolution images. It includes 248,015 manually annotated instances, collected from diverse orchards in Michigan and California during the 2022 and 2023 growing seasons. The dataset used in this experiment covers three fruit categories: apples, oranges, lemons, with each image annotated using precise bounding boxes and class labels. MetaFruit poses significant challenges due to factors such as varying lighting conditions, occlusions, and clustered fruit appearances, making it well-suited for evaluating the proposed VLLFL framework in realistic agricultural scenarios. Notably, MetaFruit is the largest publicly available multi-class fruit dataset to date, surpassing prior datasets by more than an order of magnitude in the number of labeled instances.

To further evaluate the generalization capability of VLLFL, we also employ the Harmful Animal in the Field dataset \cite{harmfulanimalinthefield_dataset}. This dataset contains 9,849 images aimed at detecting animals that pose potential threats to agricultural environments. It comprises 31 distinct classes, including buffalo, chicken, deer, elephant, goat, monkey, pig, and others. In our experiments, we randomly select 4 classes for evaluation to simulate varied detection tasks.

Both datasets are partitioned into training, validation, and test sets using an 8:1:1 ratio.

\subsubsection{Training parameter}
For few-shot training, the batch size is set to 4. The learning rate is configured as 0.05 with a weight decay of $1 \times 10^{-4}$. The model is trained for 500 communication rounds using the AdamW optimizer. The loss function used for training is defined as follows:
\begin{equation}
\mathcal{L} = \mathcal{L}_1 + \mathcal{L}_{\mathrm{GIOU}} + \mathcal{L}_{\mathrm{Cons}},
\end{equation}

The loss function used in our framework is composed of three components:

\begin{itemize}
    \item $\mathcal{L}_1$ denotes the mean absolute error (MAE), which measures the absolute difference between the predicted bounding box parameters and their ground-truth counterparts. While $\mathcal{L}_1$ ensures numerical closeness in coordinate predictions, it does not directly evaluate spatial overlap.

    \item To compensate for this limitation, $\mathcal{L}_{\mathrm{GIOU}}$ (Generalized Intersection over Union) is employed to assess the degree of overlap between predicted and ground-truth bounding boxes.

    \item Additionally, $\mathcal{L}_{\mathrm{Cons}}$ is incorporated to enhance consistency between predicted object categories and the corresponding language tokens, effectively refining class classification performance.

\end{itemize}

During inference, a bounding box confidence threshold of 0.3 and a text similarity threshold of 0.25 are applied for post-processing. The prompt width is set to 4, meaning that four prompt vectors are generated per class. In each communication round, 70\% of clients are randomly selected to participate in the federated learning process.

All training and evaluation are conducted on a server running Ubuntu 20.04, equipped with an AMD Ryzen Threadripper PRO 3955WX 16-core CPU and an NVIDIA RTX A6000 GPU.

\subsubsection{Baseline method}
In our evaluation, we adopt Federated Context Optimization (FedCoOp) \cite{guo2023promptfl} as the baseline method to highlight the advantages of the proposed VLLFL framework. FedCoOp learns continuous prompt embeddings to enhance the performance of vision-language models (VLMs) across federated tasks. Once optimized, these prompt embeddings remain fixed during inference. This is a key difference between FedCoOp and our proposed approach. In contrast, VLLFL is designed to train a context-aware prompt generator capable of dynamically producing optimized prompts for different classes, rather than relying on static, pre-trained embeddings. This flexibility allows VLLFL to adapt to diverse task distributions across clients and achieve improved generalization in federated settings.

\subsection{Results}
\subsubsection{Model training}

\begin{figure}[htp]
    \centering
    \includegraphics[width=0.75\linewidth]{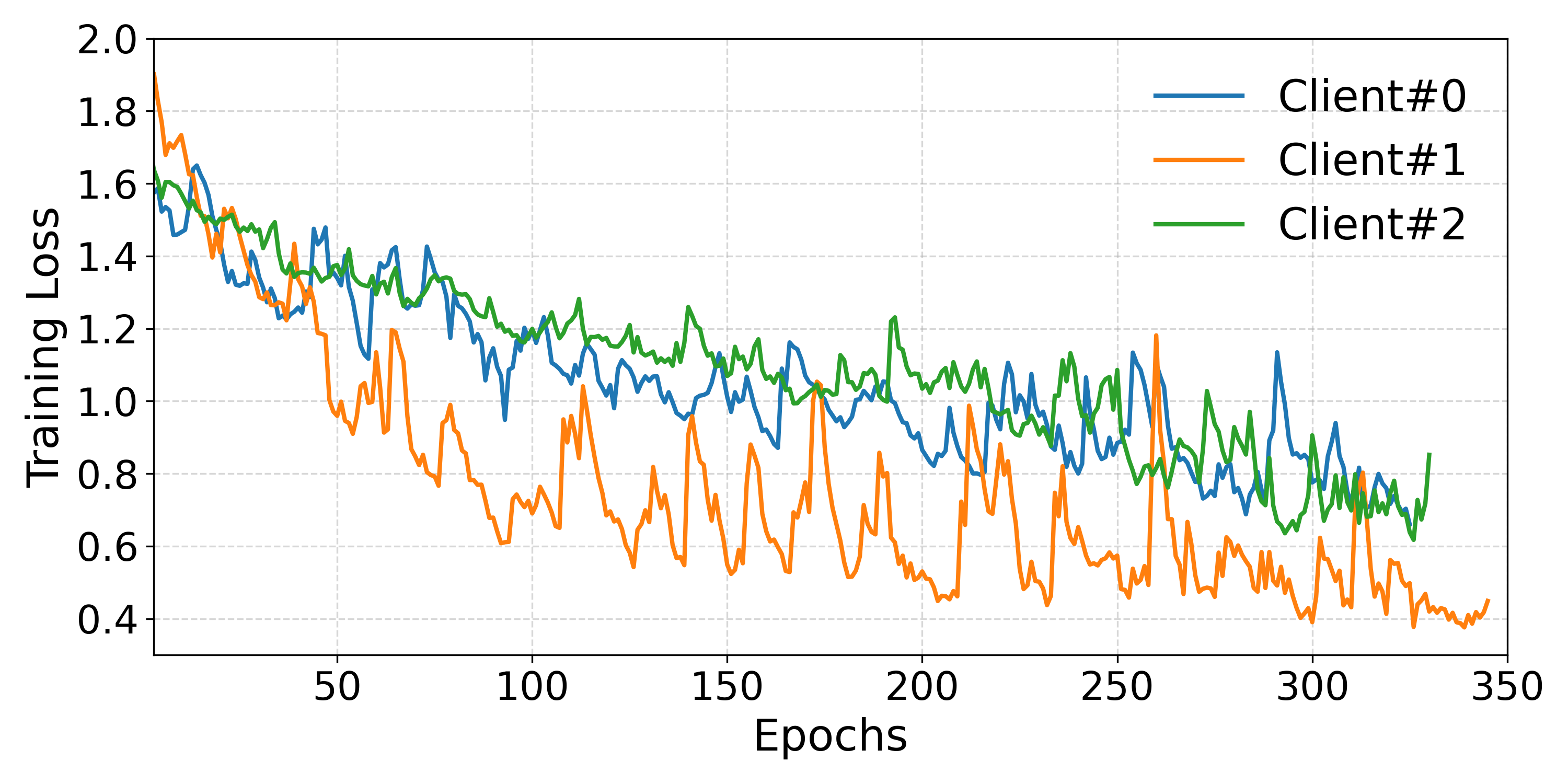}
    \caption{Training loss over training epochs}
    \label{fig:loss}
\end{figure}

In the proposed VLLFL framework, training is collaboratively conducted by the clients and the central server. In our experiments, we simulate a federated learning scenario with three clients, each representing a distinct farmland. During each training epoch, two clients are randomly selected to participate in the model update. These selected clients upload the weights of their locally trained prompt generators to the server for aggregation and global model refinement. The training process spans a total of 500 communication rounds, with each client participating in approximately 325 rounds on average. Due to the lightweight nature of the prompt generator, only a small dataset is required for effective training. The training dynamics, including the loss trajectories of all three clients over the course of federated training, are illustrated in Fig. \ref{fig:loss}.  

As shown in Fig. \ref{fig:loss}, the training losses for all clients consistently decrease over time, indicating effective learning across the federated training process. As training progresses, each local client improves its ability to detect the specific object classes relevant to its task. Notably, Client 1 exhibits a faster loss reduction compared to the other two clients. This is likely due to the nature of its primary target—oranges—which have distinct and easily identifiable visual features, such as color, making them easier to detect. 

Despite the overall downward trend, several minor fluctuations in the loss curves are observed. Such variations are expected in federated learning, as periodically aggregated model parameters from all clients are synchronized back to each participating client. This synchronization balances the knowledge across clients, ensuring that the global model generalizes well to all object classes. However, it can momentarily disrupt local convergence and introduce brief instabilities in training.

These results demonstrate the effectiveness of the VLLFL framework and suggest that approximately 325 communication rounds are sufficient for both local convergence and successful global model aggregation.

\subsubsection{VLLFL performance}
\begin{figure}[htp]
    \centering
    \includegraphics[width=0.75\linewidth]{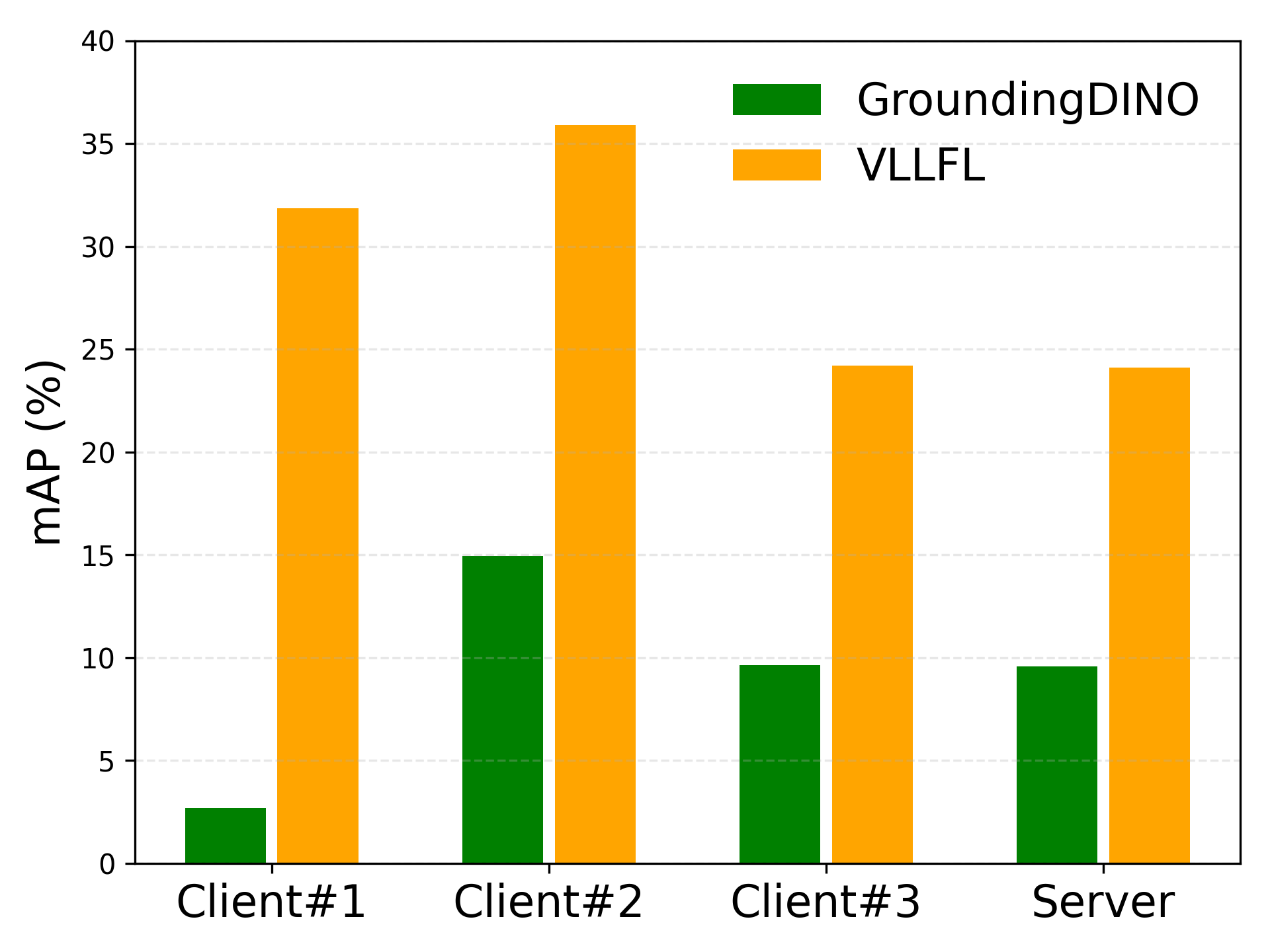}
    \caption{Detection accuracy comparison with GroundingDINO}
    \label{fig:comparison}
\end{figure}

We use the fruit detection task to evaluate the efficiency of the proposed VLLFL framework. As shown in Fig. \ref{fig:comparison}, we report the detection accuracy of each local client as well as the global model on the server. The accuracy of each client reflects its performance on detecting only its specific object classes, while the accuracy of the global model represents its ability to detect all object classes across all clients.

As illustrated in Fig. \ref{fig:comparison}, the detection accuracy of the base model, GroundingDINO, without any retraining, is significantly lower than that achieved by the proposed VLLFL framework across both local clients and the global server. For example, in Client 1, GroundingDINO achieves only 2.71\% mean Average Precision (mAP) for apple detection—the lowest among all clients. On the global server, the base model reaches just 9.59\% mAP across all object classes. Although GroundingDINO has general open-vocabulary detection capabilities due to its pretraining, it lacks domain-specific knowledge required for fine-grained classification of visually similar fruits such as apples, oranges, and lemons. This limitation results in poor detection performance without task-specific adaptation. 

In contrast, the proposed VLLFL framework achieves a global mAP of 24.12\% after federated training, more than doubling the detection accuracy of the base model. This improvement can be attributed to the context-aware prompt generator, which learns to generate class-specific prompts by aggregating semantic knowledge from all clients. These prompts guide the vision-language model to attend to more discriminative features, thereby enhancing detection performance. In local client evaluations, the VLLFL model also consistently outperforms the baseline, demonstrating its ability not only to support global model convergence across heterogeneous tasks but also to improve local model performance by sharing collective knowledge. These results validate the efficiency and effectiveness of VLLFL in enhancing vision-language model performance through federated prompt learning.

\subsubsection{Baseline Comparison}
\begin{figure}[htp]
    \centering
    \includegraphics[width=0.75\linewidth]{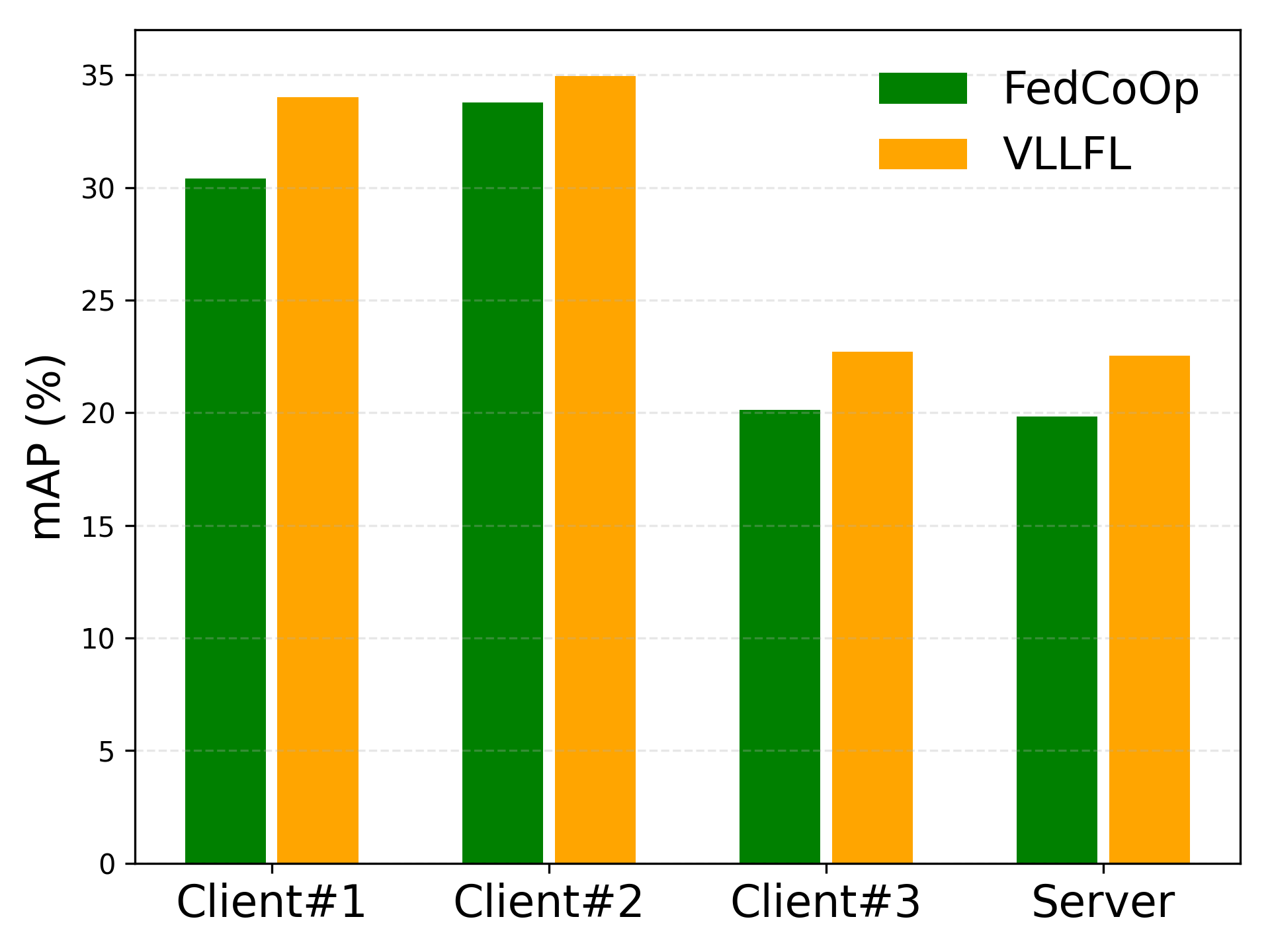}
    \caption{Detection accuracy comparison with baseline FedCoOp}
    \label{fig:baseline}
\end{figure}

To further demonstrate the advantages of the proposed VLLFL framework, we conduct a comparative evaluation against a baseline method, FedCoOp. While FedCoOp is also a federated learning approach designed for joint prompt optimization, it lacks context-awareness. Specifically, FedCoOp seeks to learn a single, task-specific prompt during training, which remains fixed at inference time regardless of the input text. In contrast, VLLFL employs a context-aware prompt generator capable of dynamically producing prompts conditioned on the input class names, allowing for better adaptability and generalization. The comparison results between VLLFL and FedCoOp are presented in Fig. \ref{fig:baseline}.

As shown in Fig. \ref{fig:baseline}, both FedCoOp and the proposed VLLFL framework improve the performance of the base GroundingDINO model, highlighting the effectiveness of text prompts in enhancing VLM performance. Text prompts provide additional semantic cues that help the model perform better on tasks where it lacks domain-specific knowledge. In our case, the pre-trained GroundingDINO is a general-purpose base model that has not been fine-tuned for fruit detection in agricultural environments. Thus, optimized prompts serve to bridge this knowledge gap and guide the model toward accurate object detection.

Notably, VLLFL consistently outperforms FedCoOp across both local clients and the global server. Unlike FedCoOp, which learns a fixed task-specific prompt during training, VLLFL employs a context-aware prompt generator capable of dynamically adapting to different input class names. This adaptability significantly enhances the generalization capability of the framework. Federated learning involves aggregating knowledge across diverse clients and the central server, which may introduce noise or interference due to heterogeneous data distributions. VLLFL demonstrates better performance under these conditions, primarily due to its context-aware prompt generation, which allows for more flexible and personalized representations across tasks.

These results validate not only the general utility of text prompts in improving VLM performance, but also the superior effectiveness of the VLLFL framework, driven by its ability to generate adaptive, context-aware prompts.

\subsubsection{Performance on Fine-tune model}

In the previous experiment, we demonstrated that the text prompt generator significantly improves detection accuracy, especially when the base model lacks prior knowledge about the task. The original GroundingDINO model used in our setup had not been fine-tuned for fruit detection in agricultural contexts. In such cases, the prompt generator effectively injects new semantic information into the model, guiding it toward correct predictions. 

\begin{figure}[htp]
    \centering
    \includegraphics[width=0.75\linewidth]{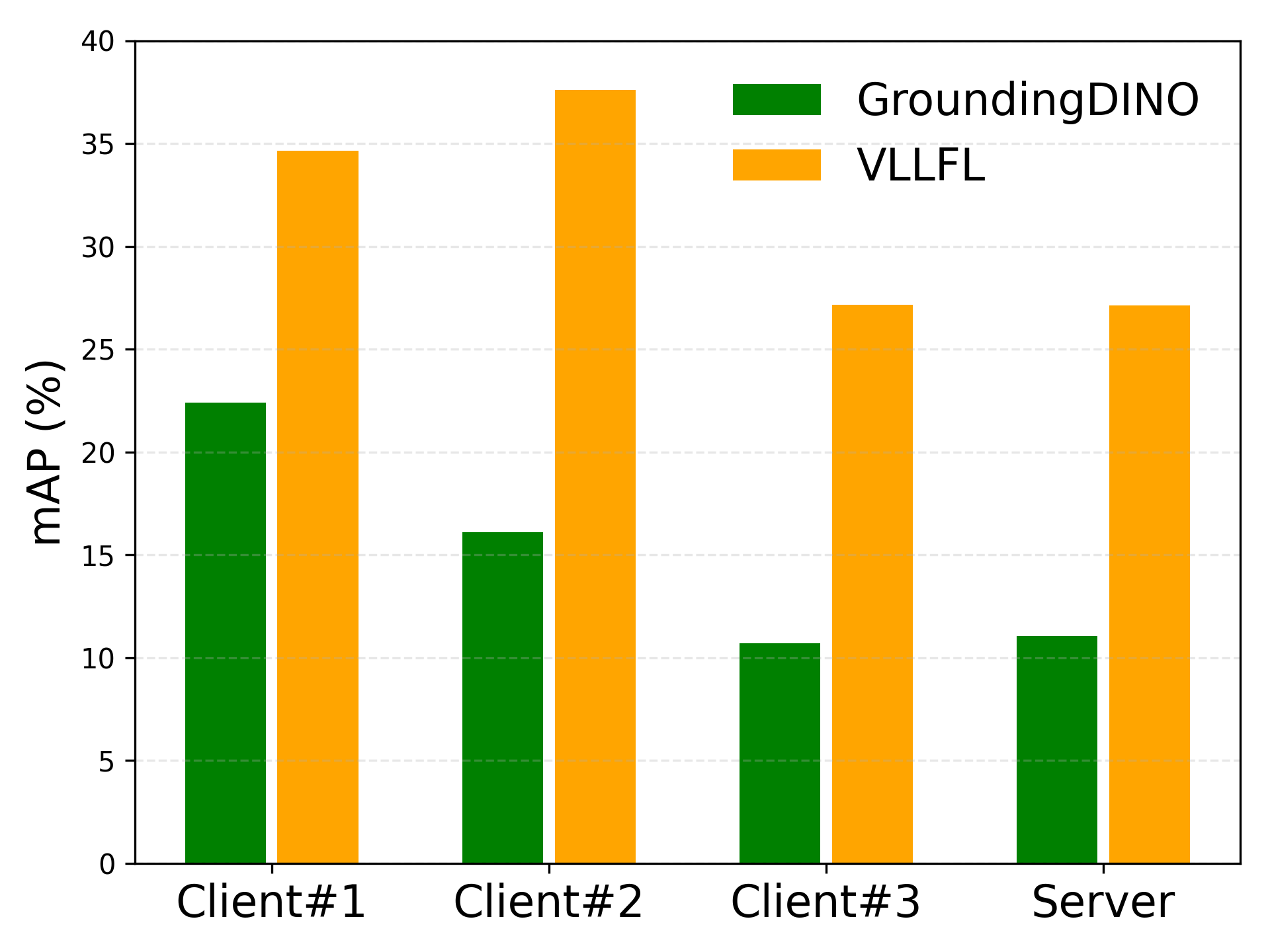}
    \caption{Performance based on fine-tuned GroundingDINO}
    \label{fig:fine-tune}
\end{figure}

\begin{figure}[htp]
    \centering
    \includegraphics[width=0.75\linewidth]{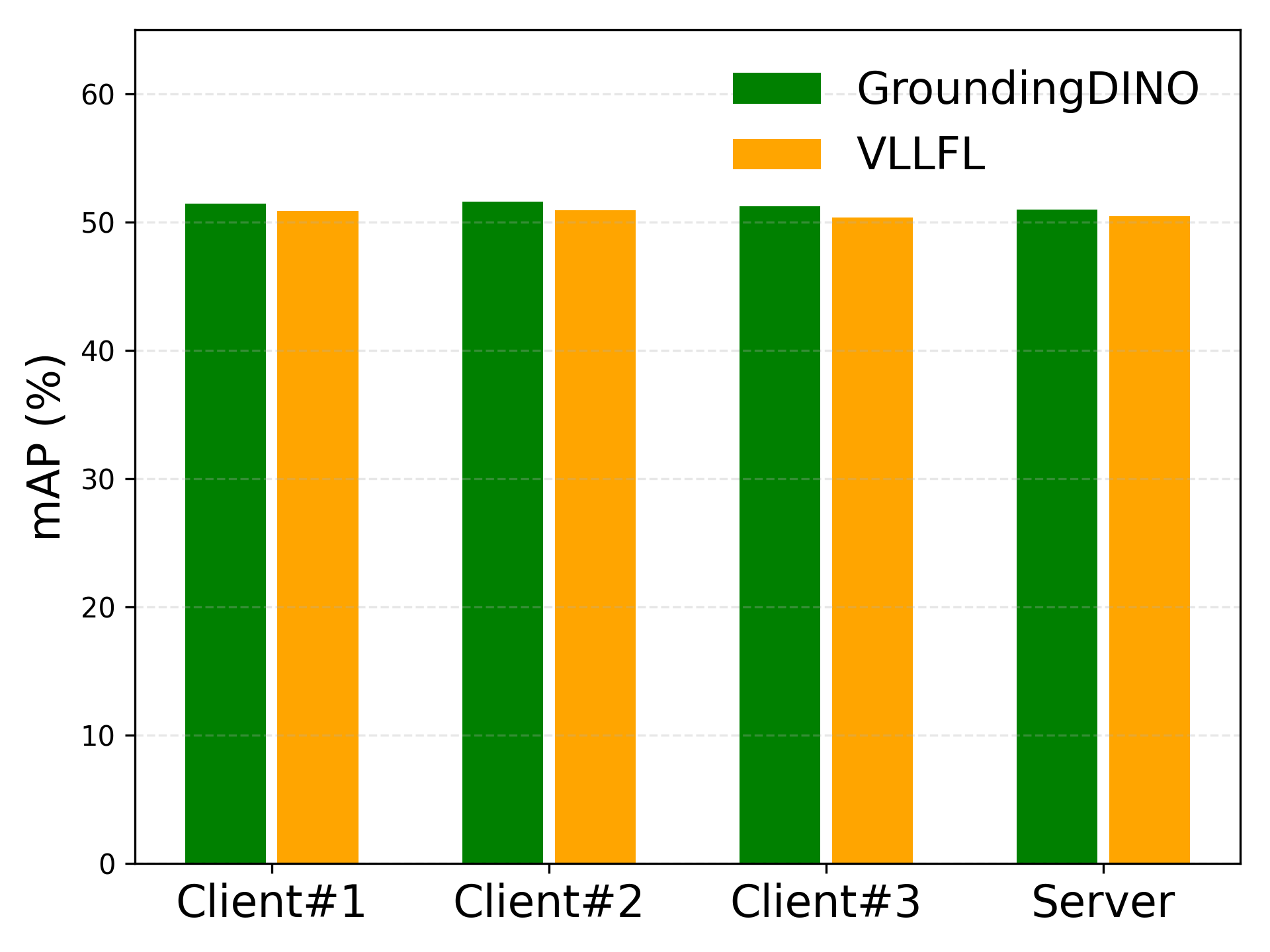}
    \caption{Performance based on fully optimized GroundingDINO}
    \label{fig:fully-tune}
\end{figure}

However, the benefits provided by VLLFL may differ when the base model already possesses a certain level of task-specific knowledge. To explore this, we conduct an additional experiment where we fine-tune the base GroundingDINO model using a few-shot approach. Specifically, we randomly select four images per class for fine-tuning. We then evaluate the performance of VLLFL based on this fine-tuned model. The comparative results for the fine-tuned base model and the fully optimized VLLFL framework are presented in Fig. \ref{fig:fine-tune} and Fig. \ref{fig:fully-tune}, respectively.

Fig. \ref{fig:fine-tune} illustrates the detection accuracy of the fine-tuned GroundingDINO model and the VLLFL framework trained on top of this fine-tuned base. The few-shot fine-tuning (conducted over two rounds) leads to a noticeable improvement in performance, as the base model acquires task-specific knowledge that helps it make more accurate predictions for fruit detection. Despite this enhancement, the proposed VLLFL framework continues to provide additional performance gains. This is because the prompt generator introduces valuable semantic cues from a text-driven perspective. Rather than learning implicit correlations between textual and visual features during model training—as traditional VLMs do—VLLFL explicitly reshapes the input prompts to better align with specific visual characteristics in the image. This design allows VLLFL to directly and efficiently enhance detection performance, whether the base model is fine-tuned or not, thereby demonstrating its general applicability and robustness.

However, the results for a fully fine-tuned base model present a different outcome, as shown in Fig. \ref{fig:fully-tune}. In this experiment, we fine-tuned the GroundingDINO model to its maximum achievable accuracy. Under this condition, the VLLFL framework does not yield further improvements in detection accuracy. This is because the base model, having already acquired comprehensive knowledge about the fruit detection task through full fine-tuning, no longer benefits from additional semantic guidance. The prompt generator is unable to introduce new or complementary information that the base model has not already learned. As a result, VLLFL achieves accuracy levels comparable to the fully fine-tuned GroundingDINO model, indicating that its advantages are most prominent when the base model lacks domain-specific expertise.

As previously discussed, centralized dataset collection raises significant privacy concerns, particularly when handling sensitive agricultural data across distributed farms. The proposed VLLFL framework addresses this challenge by enhancing the performance of VLM through a lightweight, federated learning approach that avoids the need for raw data sharing. This design allows for the deployment of large VLMs in agricultural applications while preserving data privacy. The experimental results validate the effectiveness of VLLFL in improving VLM performance under privacy-preserving conditions. Although the performance ceiling is ultimately constrained by the maximum achievable accuracy of the base model, VLLFL offers a promising solution for deploying vision-language models in agriculture. It enables performance improvements through a compact, context-aware prompt generator, making it feasible to run on lightweight systems without compromising data confidentiality.

\subsubsection{Number of participants}

\begin{figure}[htp]
    \centering
    \includegraphics[width=0.75\linewidth]{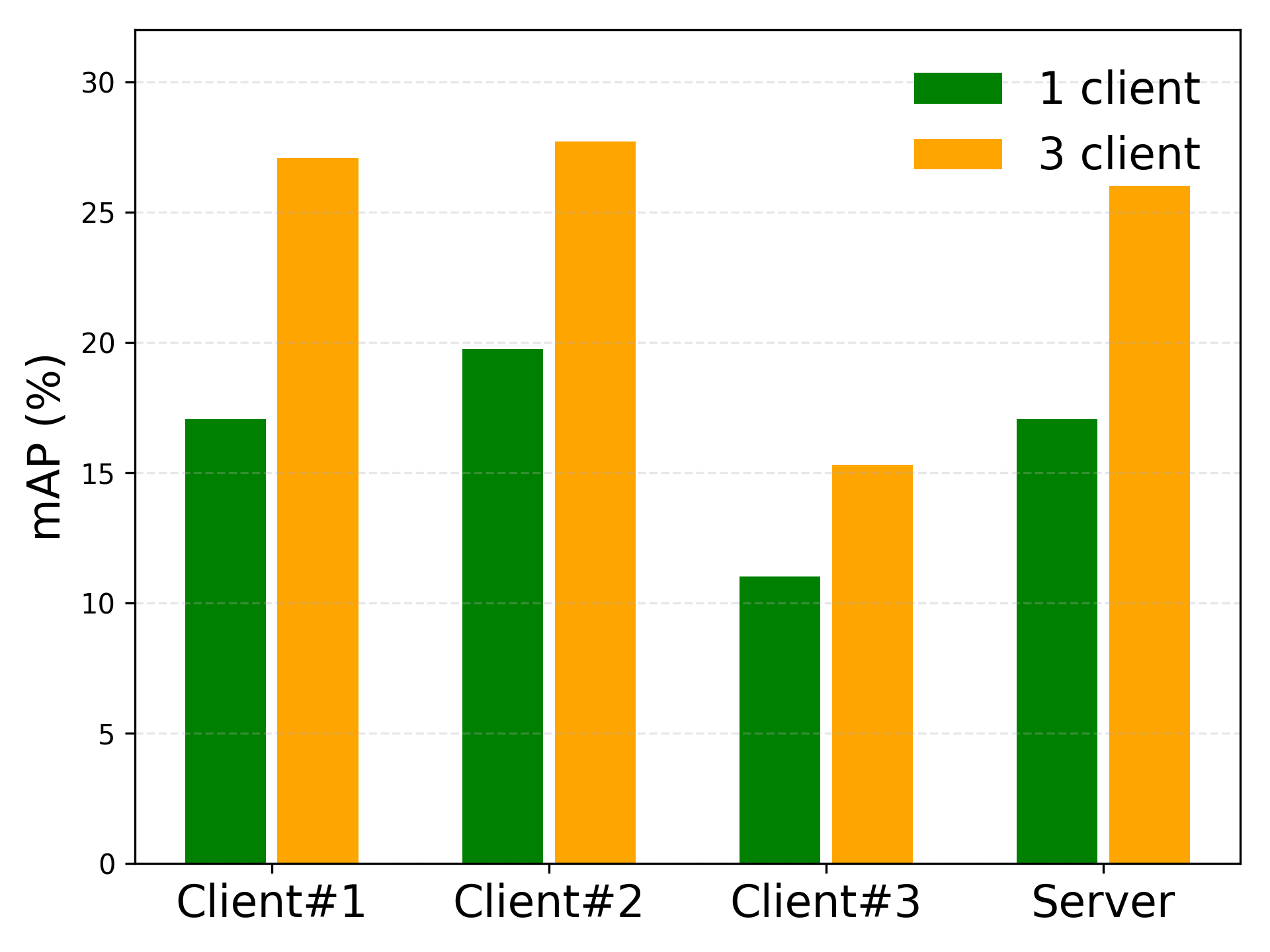}
    \caption{Performance comparison with different participant rate}
    \label{fig:participants}
\end{figure}

In federated learning, the participation rate—defined as the proportion of clients involved in each training round—plays a critical role in shaping overall model performance. A higher participation rate typically leads to faster convergence and improved stability during training. Fig. \ref{fig:participants} presents the results of training the VLLFL framework over 100 communication rounds under varying participation rates, illustrating the impact of client participation on model performance.

As shown in Fig. \ref{fig:participants}, increasing the participation rate from one to three clients per round results in a noticeable improvement in mAP across all clients and the global server. These results indicate that involving more clients per round facilitates faster convergence, as the VLLFL framework benefits from access to a more diverse set of training data during each communication round. However, it is important to note that, given sufficient training time, VLLFL trained with lower participation rates can still achieve comparable performance. The primary advantage of a higher participation rate lies in the acceleration of the learning process rather than the final model accuracy. That said, involving more clients per round also increases communication overhead, as a greater number of local updates must be transmitted and aggregated during each round. Therefore, while a higher participation rate improves convergence speed and early-stage performance, it also demands greater communication resources—highlighting the need to balance efficiency, accuracy, and resource constraints in practical federated learning deployments.

\subsubsection{Generalization to different task}

To evaluate the generalization capability of the proposed VLLFL framework, we applied it to a distinct agricultural object detection task using the Harmful Animal Detection Dataset.

\begin{figure}[htp]
    \centering
    \includegraphics[width=0.75\linewidth]{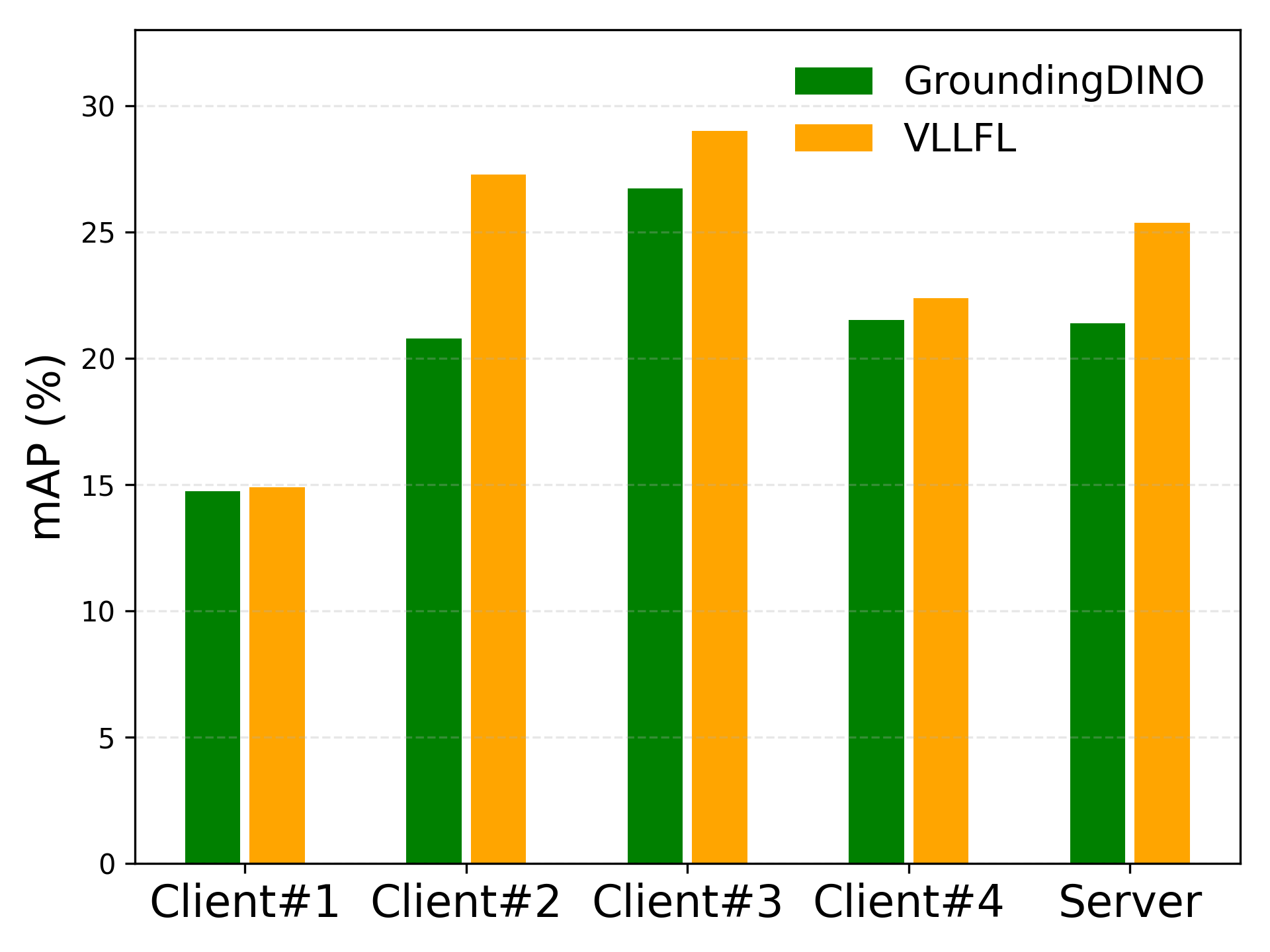}
    \caption{Detection performance on harmful animal detection task}
    \label{fig:animal}
\end{figure}

The results, illustrated in Fig. \ref{fig:animal}, show a clear performance advantage of the proposed VLLFL framework over the baseline model, GroundingDINO. Across all clients and the global server, VLLFL consistently outperforms the base model, demonstrating its ability to generalize effectively to different agricultural object detection tasks. The consistent performance of VLLFL across heterogeneous tasks highlights its robustness and scalability. These characteristics make it a promising solution for a wide range of agricultural applications, enabling privacy-preserving training while improving efficiency and model performance.

\section{Conclusion}
In this paper, we introduced VLLFL, a vision-language model-based lightweight federated learning framework for enhancing object detection in smart agriculture while preserving data privacy. 
By training and exchanging only a compact prompt generator on local devices, VLLFL achieves a 99.3\% reduction in communication overhead while enhancing the performance of large vision-language models, such as GroundingDINO, across a range of agricultural detection tasks. Experimental results demonstrated consistent and substantial improvements in detection accuracy, confirming the framework’s scalability, robustness, and suitability for real-world deployment in privacy-sensitive agricultural environments.

The performance of VLLFL is currently constrained by the base VLM’s knowledge, as it relies solely on text prompts. In future work, we plan to incorporate both textual and visual feature prompts to further enhance VLM performance, even when the model already possesses substantial knowledge of the target task.

\printbibliography

\end{document}